\documentclass[11pt,a4paper]{article}
\usepackage[hyperref]{emnlp-ijcnlp-2019}
\usepackage{times}
\usepackage{latexsym}
\usepackage{multirow}

\usepackage{url}

\aclfinalcopy % Uncomment this line for the final submission

\title{Automatically Inferring Gender Associations from Language}

\author{Serina Chang \\
  Department of Computer Science \\
  Columbia University\thanks{\hspace{.2cm}Since writing this paper, Serina Chang has moved to the Department of Computer Science at Stanford University.} \\
  {\tt serinac@stanford.edu} \\\And
  Kathleen McKeown \\
  Department of Computer Science \\
  Columbia University \\
  {\tt kathy@cs.columbia.edu} \\}

\date{}

\begin{document}
\maketitle
\begin{abstract}
In this paper, we pose the question: do people talk about women and men in different ways? We introduce two datasets and a novel integration of approaches for automatically inferring gender associations from language, discovering coherent word clusters, and labeling the clusters for the semantic concepts they represent. The datasets allow us to compare how people write about women and men in two different settings –- one set draws from celebrity news and the other from student reviews of computer science professors. We  demonstrate that there \textit{are} large-scale differences in the ways that people talk about women and men and that these differences vary across domains. Human evaluations show that our methods significantly outperform strong baselines. 
\end{abstract}

\section{Introduction}

It is well-established that gender bias exists in language -- for example, we see evidence of this given the prevalence of sexism in abusive language datasets \cite{waseem2016, jha2017}. However, these are extreme cases of gender norms in language, and only encompass a small proportion of speakers or texts.

Less studied in NLP is how gender norms manifest in everyday language -- do people talk about women and men in different ways? These types of differences are far subtler than abusive language, but they can provide valuable insight into the roots of more extreme acts of discrimination. Subtle differences are difficult to observe because each case on its own could be attributed to circumstance, a passing comment or an accidental word. However, at the level of hundreds of thousands of data points, these patterns, if they do exist, become undeniable. Thus, in this work, we introduce new datasets and methods so that we can study subtle gender associations in language at the large-scale.

Our contributions include:
{\begin{itemize}
    \item Two datasets for studying language and gender, each consisting of over 300K sentences.
    \item Methods to infer gender-associated words and labeled clusters in any domain.
    \item Novel findings that demonstrate in both domains that people do talk about women and men in different ways.
\end{itemize}}

Each contribution brings us closer to modeling how gender associations appear in everyday language. In the remainder of the paper, we present related work, our data collection, methods and findings, and human evaluations of our system.\footnote{Our datasets and code are available at cs.columbia.edu/nlp/tools.cgi\#gendered\%20corpus and github.com/serinachang5/gender-associations, respectively.}

\section{Related Work}
The study of gender and language has a rich history in social science. Its roots are often attributed to Robin Lakoff, who argued that language is fundamental to gender inequality, ``reflected in both the ways women are expected to speak, and the ways in which women are spoken of'' \cite{lakoff1973}. Prominent scholars following Lakoff have included Deborah Tannen \citeyearpar{tannen1990}, Mary Bucholtz and Kira Hall \citeyearpar{bucholtz1995}, Janet Holmes \citeyearpar{holmes2003}, Penelope Eckert \citeyearpar{eckert2003}, and Deborah Cameron \citeyearpar{cameron2008}, along with many others.

In recent decades, the study of gender and language has also attracted computational researchers. Echoing Lakoff's original claim, a popular strand of computational work focuses on differences in how women and men talk, analyzing key lexical traits \cite{boulis2005, argamon2007, bamman2014} and predicting a person's gender from some text they have written \cite{rao2010, jurgens2017}. There is also research studying how people talk \textit{to} women and men \cite{voigt2018}, as well as how people talk \textit{about} women and men, typically in specific domains such as sports journalism \cite{fu2016}, fiction writing \cite{fast2016}, movie scripts \cite{sap2017}, and Wikipedia biographies \cite{wagner2015, wagner2016}. Our work builds on this body by diving into two novel domains: celebrity news, which explores gender in pop culture, and student reviews of CS professors, which examines gender in academia and, particularly, the historically male-dominated field of CS. Furthermore, many of these works rely on manually constructed lexicons or topics to pinpoint gendered language, but our methods automatically infer gender-associated words and labeled clusters, thus reducing supervision and increasing the potential to discover subtleties in the data.

Modeling gender associations in language could also be instrumental to other NLP tasks. Abusive language is often founded in sexism \cite{waseem2016, jha2017}, so models of gender associations could help to improve detection in those cases. Gender bias also manifests in NLP pipelines: prior research has found that word embeddings preserve gender biases \cite{bolukbasi2016, caliskan2017, garg2018}, and some have developed methods to reduce this bias \cite{zhao2018, zhao2019}. Yet, the problem is far from solved; for example, \citealt{gonen2019} showed that it is still possible to recover gender bias from ``de-biased'' embeddings. These findings further motivate our research, since before we can fully reduce gender bias in embeddings, we need to develop a deeper understanding of how gender permeates through language in the first place.
%KM-final - if there is space, it might be worth pointing out that some of these approaches use a seed word list of gendered words. I remember in particular the one at NAACL with Ido Dagan as author that they did this.

We also build on methods to cluster words in word embedding space and automatically label clusters. Clustering word embeddings has proven useful for discovering salient patterns in text corpora \cite{wilson2018, demszky2019}. Once clusters are derived, we would like them to be interpretable. Much research simply considers the top-\textit{n} words from each cluster, but this method can be subjective and time-consuming to interpret. Thus, there are efforts to design methods of automatic cluster labeling \cite{manning2008}. We take a similar approach to \citealt{poostchi2018}, who leverage word embeddings and WordNet during labeling, and we extend their method with additional techniques and evaluations. 

\section{Data Collection}

% Our datasets draw from two domains: celebrity news and reviews of professors. These domains are particularly interesting to study, because they are both involved in shaping perceptions of gender. Celebrities are the most visible individuals in pop culture and professors are formative role models for youth, so these figures may influence how people perceive themselves and those around them.

\begin{table}
    \centering
    \begin{tabular}{|r|c|c|}
        \hline
         & \textbf{Celeb} & \textbf{Professor} \\
         \hline
        Num. texts & 15,917 & 283,973 \\
        Num. sentences & 342,645 & 976,677 \\
        Fem-Male prop. & .67 / .33 & .28 / .72 \\
        \hline
    \end{tabular}
    \caption{Summary statistics of our datasets.}
    \label{tab:data}
\end{table}

Our first dataset contains articles from celebrity magazines \textit{People}, \textit{UsWeekly}, and \textit{E!News}. We labeled each article for whether it was reporting on men, women, or neither/unknown.\footnote{Our method unfortunately could not take into account non-binary gender identities, as it relied on she/her and he/his pronouns, and could not easily integrate the singular they/them, nor could we find sufficient examples of ze/zir or other non-binary pronouns in our data. That said, we will continue striving towards better inclusion, and hope in future work to expand our methods beyond the binary.} To do this, we first extracted the article's topic tags. Some of these tags referred to people, but others to non-people entities, such as ``Gift Ideas'' or ``Health.'' To distinguish between these types of tags, we queried each tag on Wikipedia and checked whether the top page result contained a ``Born'' entry in its infobox -- if so, we concluded that the tag referred to a person.

Then, from the person's Wikipedia page, we determined their gender by checking whether the introductory paragraphs of the page contained more male or female pronouns. This method was simple but effective, since pronouns in the introduction almost always resolve to the subject of that page. In fact, on a sample of 80 tags that we manually annotated, we found that comparing pronoun counts predicted gender with perfect accuracy. Finally, if an article tagged at least one woman and did not tag any men, we labeled the article as Female; in the opposite case, we labeled it as Male.

Our second dataset contains reviews from RateMyProfessors (RMP), an online platform where students can review their professors. We included all 5,604 U.S. schools on RMP, and collected all reviews for CS professors at those schools. We labeled each review with the gender of the professor whom it was about, which we determined by comparing the count of male versus female pronouns over all reviews for that professor. This method was again effective, because the reviews are expressly written about a certain professor, so the pronouns typically resolve to that professor.

In addition to extracting the text of the articles or reviews, for each dataset we also collected various useful metadata. For the celebrity dataset, we recorded each article's timestamp and the name of the author, if available. Storing author names creates the potential to examine the relationship between the gender of the author and the gender of the subject, such as asking if there are differences between how women write about men and how men write about men. In this work, we did not yet pursue this direction because we wanted to begin with a simpler question of how gender is discussed: regardless of the gender of the authors, what is the content being put forth and consumed? Furthermore, we were unable to extract author gender in the professor dataset since the RMP reviews are anonymous. However, in future work, we may explore the influence of author gender in the celebrity dataset.

For the professor dataset, we captured metadata such as each review's rating, which indicates how the student feels about the professor on a scale of AWFUL to AWESOME. This additional variable in our data creates the option in future work to factor in sentiment; for example, we could study whether there are differences in language used when criticizing a female versus a male professor.

\section{Inferring Word-Level Associations}

Our first goal was to discover words that are significantly associated with men or women in a given domain. We employed an approach used by \citealt{bamman2014} in their work to analyze differences in how men and women write on Twitter.

\subsection{Methods}
First, to operationalize, we say that term $i$ is \textit{associated} with gender $j$ if, when discussing individuals of gender $j$, $i$ is used with unusual frequency -- which we can check with statistical hypothesis tests. Let $f_i$ represent the likelihood of $i$ appearing when discussing women or men. $f_i$ is unknown, but we can model the distribution of all possible $f_i$ using the corpus of texts that we have from the domain. We construct a gender-balanced version of the corpus by randomly undersampling the more prevalent gender until the proportions of each gender are equal. Assuming a non-informative prior distribution on $f_i$, the posterior distribution is Beta($k_i$, $N - k_i$), where $k_i$ is the count of $i$ in the gender-balanced corpus and $N$ is the total count of words in that corpus.

As \citealt{bamman2014} discuss, ``the distribution of the gender-specific counts can be described by an integral over all possible $f_i$. This integral defines the Beta-Binomial distribution \cite{gelman2004}, and has a closed form solution.'' We say that term $i$ is significantly associated with gender $j$ if the cumulative distribution at $k_{ij}$ (the count of $i$ in the $j$ portion of the gender-balanced corpus) is $p \leq 0.05$. As in the original work, we apply the Bonferroni correction \cite{dunn1961} for multiple comparisons because we are computing statistical tests for thousands of hypotheses.

\subsection{Findings}

\begin{table}
    \centering
    \begin{tabular}{|p{3.5cm}|p{3.5cm}|}
         \hline
          \textbf{Female-Associated} & \textbf{Male-Associated} \\
          \hline 
          girl, cover, husband, wedding, gown, fashion, mom, pregnancy, photo, top, hair, look & movie, president, wife, dad, death, film, host, assault, claim, misconduct, action, director \\
         \hline
         respond, email, recommend, help, love, accept, need, send, reply, communicate & learn, teach, know, write, lecture, challenge, solve, ramble, push, joke, bore \\
         \hline
         easy, rude, wonderful, kind, caring, hot, strict, timely, mean, disorganized, beautiful & knowledgeable, real, challenging, brilliant, arrogant, hard, passionate, practical \\
         \hline
    \end{tabular}
    \caption{\textit{Top}: Sample from the top-25 most gender-associated \textbf{nouns} in the celebrity domain. \textit{Middle}: professor domain, sample from top-25 \textbf{verbs}. \textit{Bottom}: professor domain, sample from top-25 \textbf{adjectives}. All associations listed are $p \leq 0.05$, with Bonferroni correction. See Appendix for all top-25 nouns, verbs, and adjectives for both genders in both domains.}
    \label{tab:word-level-results}
\end{table}

We applied this method to discover gender-associated words in both domains. In Table \ref{tab:word-level-results}, we present a sample of the most gender-associated nouns from the celebrity domain. Several themes emerge: for example, female celebrities seem to be more associated with appearance (``gown,'' ``photo,'' ``hair,'' ``look''), while male celebrities are more associated with creating content (``movie,'' ``film,'' ``host,'' ``director''). This echoes real-world trends: for instance, on the red carpet, actresses tend to be asked more questions about their appearance –- what brands they are wearing, how long it took to get ready, etc. –- while actors are asked questions about their careers and creative processes (as an example, see \citealt{selby2014}).

Table \ref{tab:word-level-results} also includes some of the most gender-associated verbs and adjectives from the professor domain. Female CS professors seem to be praised for being communicative and personal with students (``respond,'' ``communicate,'' ``kind,'' ``caring''), while male CS professors are recognized for being knowledgeable and challenging the students (``teach,'', ``challenge,'' ``brilliant,'' ``practical''). These trends are well-supported by social science literature, which has found that female teachers are praised for ``personalizing'' instruction and interacting extensively with students, while male teachers are praised for using ``teacher as expert'' styles that showcase mastery of material \cite{statham1991}.

These findings establish that there are clear differences in how people talk about women and men -- even with Bonferroni correction, there are still over 500 significantly gender-associated nouns, verbs, and adjectives in the celebrity domain and over 200 in the professor domain. Furthermore, the results in both domains align with prior studies and real world trends, which validates that our methods can capture meaningful patterns and innovatively provide evidence at the large-scale. This analysis also hints that it can be helpful to abstract from words to topics to recognize higher-level patterns of gender associations, which motivates our next section on clustering.

\section{Clustering \& Cluster Labeling}
With word-level associations in hand, our next goals were to discover coherent clusters among the words and to automatically label those clusters.

\subsection{Methods}
First, we trained domain-specific word embeddings using the Word2Vec \cite{mikolov2013} CBOW model ($w \in R^{100}$). Then, we used k-means clustering to cluster the embeddings of the gender-associated words. Since k-means may converge at local optima, we ran the algorithm 50 times and kept the model with the lowest sum of squared errors.

\begin{table*}
    \centering
    \begin{tabular}{|p{1.5cm}|p{6cm}|p{1cm}|p{6cm}|}
        \hline
          \textbf{Domain} & \textbf{Sample Words in Cluster} & \textbf{F:M} & \textbf{Top 3 Pred. Cluster Labels}\\
          \hline
          \multirow{3}{*}{Celeb} 
        %   & photo, post, pic, snap, selfie, image & 10:2 & \textit{photograph, representation, picture} \\
          & gown, top, dress, pant, skirt, neckline  & 25:0 & \textit{covering,      cloth\_covering, clothing} \\ 
          & film, release, role, character, project & 4:16 & \textit{movie, show, event} \\  
          & boyfriend, beau, hubby, wife, girlfriend & 15:7 & \textit{lover, person, relative} \\
          \hline
          \multirow{3}{*}{Professor} & response, email, contact, answer & 13:0 & \textit{statement, message, communication} \\ 
          & material, concept, topic, stuff, subject & 1:8 & \textit{content, idea, cognition} \\ 
          & teacher, woman, lady, prof, guy, dude & 5:7 & \textit{man, adult, woman} \\
          \hline
    \end{tabular}
    \caption{Sample of our clusters and predicted cluster labels. We include in the Appendix a more comprehensive table of our results. \textbf{F:M} refers to the ratio of female-associated to male-associated words in the cluster.}
    \label{tab:cluster-label-results}
\end{table*}

To automatically label the clusters, we combined the grounded knowledge of WordNet \cite{miller1995} and context-sensitive strengths of domain-specific word embeddings. Our algorithm is similar to \citealt{poostchi2018}'s approach, but we extend their method by introducing domain-specific word embeddings for clustering as well as a new technique for sense disambiguation. Given a cluster, our algorithm proceeds with the following three steps:

\begin{enumerate}
    \item \textbf{Sense disambiguation}:	The goal is to assign each cluster word to one of its WordNet synsets; let $S$ represent the collection of chosen synsets. We know that these words have been clustered in domain-specific embedding space, which means that in the context of the domain, these words are very close semantically. Thus, we choose $S^*$ that minimizes the total distance between its synsets.
    % \footnote{It is extremely computationally expensive to try all possible $S$, so in practice, we rabonfndomly divide the synset candidates into chunks of five and find the optimal sense assignment over the chunk, then we concatenate the optimal assignments from all of the chunks. This is reasonable, since the context of four other similar words should be enough to sense disambiguate. To be extra careful, we repeat the chunking process for three trials so that each time, each word is disambiguated with a new random chunk of other words. After the trials, each word is matched to a set of three senses (possibly all the same sense), and we take the majority sense for each word as its final sense assignment.}
    \item \textbf{Candidate label generation}: In this step, we generate $L$, the set of possible cluster labels. Our approach is simple: we take the union of all hypernyms of the synsets in $S^*$.
    \item \textbf{Candidate label ranking}: Here, we rank the synsets in $L$. We want labels that are as close to all of the synsets in $S^*$ as possible; thus, we score the candidate labels by the sum of their distances to each synset in $S^*$ and we rank them from least to most distance.
\end{enumerate}
In steps 1 and 3, we use WordNet pathwise distance, but we encourage the exploration of other distance representations as well.

\subsection{Findings}
Table \ref{tab:cluster-label-results} displays a sample of our results -- we find that the clusters are coherent in context and the labels seem reasonable. In the next section, we discuss human evaluations that we conducted to more rigorously evaluate the output, but first we discuss the value of these methods toward analysis.

At the word-level, we hypothesized that in the celebrity domain, women were more associated with appearance and men with creating content. Now, we can validate those hypotheses against labeled clusters -- indeed, there is a cluster labeled \textit{clothing} that is 100\% female (i.e. 100\% words are female-associated), and a 80\% male cluster labeled \textit{movie}. Likewise, in the professor domain, we had guessed that women are associated with communication and men with knowledge, and there is a 100\% female cluster labeled \textit{communication} and a 89\% male cluster labeled \textit{cognition}. Thus, cluster labeling proves to be very effective at pulling out the patterns that we believed we saw at the word-level, but could not formally validate. 

The clusters we mentioned so far all lean heavily toward one gender association or the other, but some clusters are interesting precisely because they do not lean heavily -- this allows us to see where semantic groupings do not align exactly with gender association. For example, in the celebrity domain, there is a cluster labeled \textit{lover} that has a mix of female-associated words (``boyfriend,'' ``beau,'' ``hubby'') and male-associated words (``wife,'' ``girlfriend''). Jointly leveraging cluster labels and gender associations allows us to see that in the semantic context of having a \textit{lover}, women are typically associated with male figures and men with female figures, which reflects heteronormativity in society.

\section{Human Evaluations}
To test our clusters, we employed the \textit{Word Intrusion} task \cite{chang2009}. We present the annotator with five words -- four drawn from one cluster and one drawn randomly from the domain vocabulary -- and we ask them to pick out the intruder. The intuition is that if the cluster is coherent, then an observer should be able to identify the out-of-cluster word as the intruder. For both domains, we report results on all clusters and on the top 8, ranked by ascending normalized sum of squared errors, which can be seen as a prediction of coherence. In the celebrity domain, annotators identified the out-of-cluster word 73\% of the time in the top-8 and 53\% overall. In the professor domain, annotators identified it 60\% of the time in the top-8 and 49\% overall. As expected, top-8 performance in both domains does considerably better than overall, but at all levels the precision is significantly above the random baseline of 20\%.

To test cluster labels, we present the annotator with a label and a word, and we ask them whether the word \textit{falls under} the concept. The concept is a potential cluster label and the word is either a word from that cluster or drawn randomly from the domain vocabulary. For a good label, the rate at which in-cluster words fall under the label should be much higher than the rate at which out-of-cluster words fall under. In our experiments, we tested the top 4 predicted labels and the centroid of the cluster as a strong baseline label. The centroid achieved an in-cluster rate of .60 and out-of-cluster rate of .18 (difference of .42). Our best performing predicted label achieved an in-cluster rate of .65 and an out-of-cluster rate of .04 (difference of .61), thus outperforming the centroid on both rates and increasing the gap between rates by nearly 20 points. In the Appendix, we include more detailed results on both tasks.

\section{Conclusion}
We have presented two substantial datasets and a novel integration of methods to automatically infer gender associations in language. We have demonstrated that in both datasets, there are clear differences in how people talk about women and men. Furthermore, we have shown that clustering and cluster labeling are effective at identifying higher-level patterns of gender associations, and that our methods outperform strong baselines in human evaluations. In future work, we hope to use our findings to improve performance on tasks such as abusive language detection. We also hope to delve into finer-grained analyses, exploring how language around gender interacts with other variables, such as sexual orientation or profession (e.g. actresses versus female athletes). Finally, we plan to continue widening the scope of our study -- for example, expanding our methods to include non-binary gender identities, evaluating changes in gender norms over time, and spreading to more domains, such as the political sphere.

\bibliography{emnlp2019}
\bibliographystyle{acl_natbib}
\appendix
\section{Appendix}

We include here extended results for each of the stages of work that we covered in the main paper: findings for inferring word-level gender associations (Tables \ref{tab:celeb-words-extended} and \ref{tab:prof-words-extended}), findings from clustering and cluster labeling (Tables \ref{tab:celeb-clusters-extended} and \ref{tab:prof-clusters-extended}), and results from our two evaluation tasks (Tables \ref{tab:word-intrusion} and \ref{tab:concept-word}). Tables \ref{tab:celeb-words-extended}--\ref{tab:concept-word} are on the following three pages.

\subsection{Word-Level Results}
Tables \ref{tab:celeb-words-extended} and \ref{tab:prof-words-extended} provide the top 25 most significant female-associated and male-associated nouns, verbs, and adjectives in both domains. Gender association is determined using the method described in Section 4.1.

\subsection{Clustering and Cluster Labeling Results}
Tables \ref{tab:celeb-clusters-extended} and \ref{tab:prof-clusters-extended} provide the clustering and cluster labeling results for the top-$n$ clusters in each domain, ranked by ascending normalized sum of squared errors. During clustering, we set $k$ to $\frac{N}{50}$, where $N$ was the number of embeddings being clustered. In the celebrity domain, this resulted in 45 clusters; in Table \ref{tab:celeb-clusters-extended}, we show the top 12. In the professor domain, this resulted in 16 clusters; in Table \ref{tab:prof-clusters-extended}, we show the top 6.

\subsection{Human Evaluation Results}
Table \ref{tab:word-intrusion} provides more detailed results for the Word Intrusion task. As in the main paper, we provide the results in each domain for all clusters and for just the top 8, determined by ascending normalized sum of squares.

\begin{table}[h]
    \centering
    \begin{tabular}{|c|c|c|c|c|}
        \hline
         \textbf{Domain} & \textbf{Level} & \textbf{Precision} & \textbf{Fleiss' $\kappa$} \\
         \hline
         Celeb & Top-8 & .725 & .530 \\
         Celeb & Overall & .527 & .314 \\
         Prof. & Top-8 & .600 & .216 \\
         Prof. & Overall & .488 & .212 \\
         \hline
    \end{tabular}
    \caption{Results for Word Intrusion task. All results significantly outperform the random baseline of .20 ($p \leq 0.0001$).}
    \label{tab:word-intrusion}
\end{table}

Table \ref{tab:concept-word} displays more detailed results for the cluster labeling task. As discussed in the main paper, we would like to see that the rate at which in-cluster words fall under the cluster label is much higher than the rate at which out-of-cluster words fall under. We tried the top 4 predicted labels, compared against the centroid as a baseline label. For each cluster, we tested labels against 10 in-cluster words and 3 out-of-cluster words; thus, we tested 65 questions (13 words x 5 labels) per cluster. Due to the high number of questions per cluster, we only tested 15 clusters -- the top 10 labeled in the celebrity domain and the top 5 labeled in the professor domain. We report results over all tested clusters, but broken down by label type (e.g. centroid or $2^{nd}$ predicted label).

\begin{table*}[]
    \caption{Top 25 most gender-associated nouns, verbs, and adjectives in the celebrity domain. Words are listed in order of decreasing significance, but all words fall under $p \leq$ 0.05, with Bonferroni correction.}
    \centering
    \begin{tabular}{|c|p{6cm}|p{6cm}|}
        \hline
         & \textbf{Female-Associated} & \textbf{Male-Associated} \\
         \hline
         Nouns & girl, eye, cover, star, sister, shoulder, husband, lady, baby, issue, wedding, actress, reality, daughter, carpet, gown, fashion, mom, pregnancy, photo, top, hair, look, outfit, mother & movie, actor, president, wife, dad, death, film, host, news, statement, father, girlfriend, man, assault, allegation, attorney, claim, investigation, misconduct, behavior, lawsuit, action, guy, director, harassment \\
         \hline
         Verbs & caption, wear, look, keep, love, date, match, rock, use, accessorize, style, pair, feel, share, show, shop, fit, plunge, model, apply, reveal, flaunt, open, tone, color & say, deny, accord, claim, accuse, allege, would, apologize, publish, fire, hear, play, come, continue, ask, involve, pass, replace, investigate, charge, win, rap, pay, state, sue \\
         \hline
         Adj. & red, pink, pregnant, sexy, sheer, white, black, beautiful, stylish, nude, stunning, chic, natural, blonde, new, blue, sweet, glam, loose, hot, oversized, casual, gorgeous, toned, little & sexual, consensual, inappropriate, furious, guilty, alleged, oral, presidential, late, republican, deadpool, many, financial, bad, political, public, false, russian, comic, non, dead, detailed, numerous, untrue, criminal \\
         \hline
    \end{tabular}
    \label{tab:celeb-words-extended}
\end{table*}

\begin{table*}[]
    \caption{Top 25 most gender-associated nouns, verbs, and adjectives in the professor domain. Words are listed in order of decreasing significance, but all words fall under $p \leq$ 0.05, with Bonferroni correction, aside from the last three terms listed for Female-Associated Verbs and the last four terms listed for Male-Associated Verbs.}
    \centering
    \begin{tabular}{|c|p{6cm}|p{6cm}|}
        \hline
         & \textbf{Female-Associated} & \textbf{Male-Associated} \\
         \hline
         Nouns & class, work, teacher, person, woman, assignment, email, credit, week, step, direction, attitude, instruction, lady, instructor, date, response, discussion, sweetheart, communication, husband, point, group, mail, time & material, concept, professor, exam, sense, topic, joke, math, stuff, programming, world, department, lecture, prof, note, humor, guy, story, lecturer, curve, tangent, man, dude, genius, industry \\
         \hline
         Verbs & take, would, respond, email, recommend, help, love, submit, miss, grade, treat, follow, complete, accept, hat, need, send, reply, turn, correct, communicate, check, feel, wait, finish & learn, curve, teach, lay, talk, crack, know, write, lecture, challenge, want, relate, sleep, solve, ramble, push, start, entertain, joke, put, bore, mumble, base, cover, explain \\
         \hline
         Adj. & easy, helpful, nice, rude, online, due, wonderful, extra, sweet, kind, busy, caring, hot, late, strict, timely, annoying, horrible, mean, quick, disorganized, pleasant, responsive, beautiful, lovely & good, knowledgeable, great, smart, interesting, intelligent, difficult, real, funny, hilarious, boring, cool, dry, entertaining, favorite, challenging, brilliant, arrogant, interested, hard, passionate, old, tough, practical, excellent \\
         \hline
    \end{tabular}
    \label{tab:prof-words-extended}
\end{table*}

\begin{table*}
    \caption{Top 12 clusters out of 45 overall in the celebrity domain. Predicted labels are included if applicable -- we were only able to predict labels for clusters that contained nouns, since our clustering labeling algorithm relied on the noun taxonomy in WordNet. In the \textbf{Sample Words in Cluster} column, italics indicate female-associated terms, and non-italics indicate male-associated. \textbf{F:M} refers to the ratio of female-associated to male-associated words in the cluster.}
    \label{tab:celeb-clusters-extended}
    \centering
    \begin{tabular}{|p{5.5cm}|p{1cm}|p{1.5cm}|p{5cm}|}
        \hline
          \textbf{Sample Words in Cluster} & \textbf{F:M} & \textbf{Centroid} & \textbf{Top 3 Pred. Cluster Labels}\\
          \hline
        \textit{month, day,} year, decade, hour & 2:3 & month & time period, fundamental quantity, measure \\ 
        \hline
        \textit{engagement, marriage, divorce, relationship, split, breakup} & 6:0 & breakup & state, union, separation \\ 
        \hline
        \textit{woman,} man, guy, people, someone, anyone, one, person & 1:7 & man & person, man, adult \\ 
        \hline
        \textit{photo, post, pic, snap, selfie, image, snapshot, kiss, photograph,} tribute & 10:2 & snapshot & photograph, representation, picture \\ 
        \hline
        \textit{reveal, gush, tell, explain, dish, admit,} say, recall, reply & 7:3 & explain & N/A \\ 
        \hline
        \textit{trophy,} win, category, nominate, nomination, finalist & 1:5 & finalist & collection, condition, award \\ 
        \hline
        \textit{exclusive, peek, sneak, glimpse, preview} & 5:0 & preview & look, sensing, screening \\ 
        \hline
        \textit{difference}, allegation, claim, accusation, truth, lie & 1:5 & accusation & claim, assertion, statement \\ 
        \hline
        \textit{accessorize, sneaker, toe, jewel, floral, embroider, pink, hat, veil, turtleneck, sequin, tiara, bodysuit} & 123:1 & halter & artifact, garment, covering \\ 
        \hline
        \textit{girl, sister, baby, daughter, mom, pregnancy}, dad, father, brother, son & 14:6 & dad & person, parent, mother \\ 
        \hline
        \textit{husband, boyfriend, beau, hubby, fiance}, wife, girlfriend, finacee & 15:7 & fiance & lover, person, relative \\ 
        \hline
        \textit{shoulder, gown, top, dress, pant, skirt, diamond, neckline, waist, bra} & 25:0 & blouse & covering, cloth covering, clothing \\ 
        \hline
    \end{tabular}
\end{table*}

\begin{table*}
    \caption{Top 6 clusters out of 16 overall in the professor domain. Same details as Table \ref{tab:celeb-clusters-extended} apply.}
    \label{tab:prof-clusters-extended}
    \centering
    \begin{tabular}{|p{5.5cm}|p{1cm}|p{1.5cm}|p{5cm}|}
        \hline
          \textbf{Sample Words in Cluster} & \textbf{F:M} & \textbf{Centroid} & \textbf{Top 3 Pred. Cluster Labels}\\
          \hline
        \textit{teacher, woman, lady,} professor, prof, guy, lecturer, man, dude & 5:7 & prof & man, adult, woman \\ 
        \hline
        \textit{everything,} material, concept, topic, stuff, tangent, subject, content & 1:8 & content & content, idea, cognition \\ 
        \hline
        \textit{respond, email, response, responsive, stuff, reply, contact, answer} & 13:0 & respond & statement, message, communication \\ 
        \hline
        \textit{familiar, literate, savvy,} genius, nerd, background, scientist, geek & 5:10 & geek & person, expert, anomaly \\ 
        \hline
        \textit{communicate,} learn, teach, know, solve, explain, introduce, convey & 1:18 & simplify & N/A \\ 
        \hline
        \textit{compassionate, lovely, amazing, sweet, understanding, hot, nice,} passionate, funniest, smart, cool, inspirational, intelligent, likable & 21:44 & outgoing & N/A \\ 
        \hline
    \end{tabular}
\end{table*}

 \begin{table*}
    \centering
    \begin{tabular}{|c|c|c|c|c|}
        \hline
         \textbf{Label Type} & \textbf{In-Cluster Rate} & \textbf{Out-of-Cluster Rate} & \textbf{Difference} & \textbf{Fleiss' $\kappa$} \\
         \hline
         Centroid & .597 & .178 & .419 & .322 \\
         $1^{st}$ pred. label & .621 & .156 & .465 & .258 \\
         $2^{nd}$ pred. label & .540 & .200 & .340 & .256 \\
         $3^{rd}$ pred. label & \textbf{.653} & \textbf{.044} & \textbf{.609} & .353 \\
         $4^{th}$ pred. label & .452 & .111 & .341 & .261 \\
         \hline
    \end{tabular}
    \caption{Results for cluster labeling task. The $3^{rd}$ predicted label has a significantly lower out-of-cluster rate than the centroid and all the other predicted labels ($p \leq$ 0.02). The same label also slightly outperforms the centroid on the in-cluster rate, thus producing a much larger gap between rates than the centroid.}
    \label{tab:concept-word}
\end{table*}
\end{document}